\documentclass[conference]{IEEEtran}
\IEEEoverridecommandlockouts
\usepackage{cite}
\usepackage{amsmath,amssymb,amsfonts}
\usepackage{algorithmic}
\usepackage{graphicx}
\usepackage{textcomp}
\usepackage{xcolor}
\usepackage{multirow}
\usepackage{booktabs}  
\def\BibTeX{{\rm B\kern-.05em{\sc i\kern-.025em b}\kern-.08em
    T\kern-.1667em\lower.7ex\hbox{E}\kern-.125emX}}

\usepackage{fancyhdr}
\begin{document}

% \title{Conference Paper Title*\\

%\title{Enhancing Personality Recognition by Comparing the Predictive Power of the Hierarchy of the Big Five: Traits, Facets, and Nuances}

\title{Enhancing Personality Recognition by Comparing the Predictive Power of Traits, Facets, and Nuances}

%\author{\IEEEauthorblockN{Amir Ansari}
%\IEEEauthorblockA{\textit{University of Barcelona}\\
%Barcelona, Spain \\
%mansaran7@alumnes.ub.edu}
%\and
%\IEEEauthorblockN{Jana Subirana}
%\IEEEauthorblockA{\textit{University of Barcelona}\\
%Barcelona, Spain \\
%jsubirna@alumnes.ub.edu}
%\and
%\IEEEauthorblockN{Bruna Silva}
%\IEEEauthorblockA{\textit{University of Barcelona}\\
%Barcelona, Spain \\
%braugusd57@alumnes.ub.edu}
%\linebreakand
%\IEEEauthorblockN{Sergio Escalera}
%\IEEEauthorblockA{\textit{University of Barcelona}\\
%Barcelona, Spain \\
%sescalera@ub.edu}
%\and
%\IEEEauthorblockN{David Gallardo-Pujol}
%\IEEEauthorblockA{\textit{University of Barcelona}\\
%Barcelona, Spain \\
%david.gallardo@ub.edu}
%\and
%\IEEEauthorblockN{Cristina Palmero}
%\IEEEauthorblockA{\textit{King's College London}\\
%London, United Kingdom \\
%cristina.palmero@kcl.ac.uk}
%}

\author{

\IEEEauthorblockN{Amir Ansari}
\IEEEauthorblockA{
\textit{Universitat de Barcelona}\\
Barcelona, Spain \\
mansaran7@alumnes.ub.edu}\\
\IEEEauthorblockN{Sergio Escalera}
\IEEEauthorblockA{\textit{Universitat de Barcelona}\\
Barcelona, Spain \\
sescalera@ub.edu}
\and
\IEEEauthorblockN{Jana Subirana}
\IEEEauthorblockA{\textit{Universitat de Barcelona}\\
Barcelona, Spain \\
jsubirna@alumnes.ub.edu}\\
\IEEEauthorblockN{David Gallardo-Pujol}
\IEEEauthorblockA{\textit{Universitat de Barcelona}\\
Barcelona, Spain \\
david.gallardo@ub.edu}
\and
\IEEEauthorblockN{Bruna Silva}
\IEEEauthorblockA{\textit{Universitat de Barcelona}\\
Barcelona, Spain \\
braugusd57@alumnes.ub.edu}\\
\IEEEauthorblockN{Cristina Palmero}
\IEEEauthorblockA{\textit{King's College London}\\
London, United Kingdom \\
cristina.palmero@kcl.ac.uk}
}

\maketitle
\thispagestyle{fancy}
\begin{abstract}
Personality is a complex, hierarchical construct typically assessed through item-level questionnaires aggregated into broad trait scores. Personality recognition models aim to infer personality traits from different sources of behavioral data. However, reliance on broad trait scores as ground truth, combined with limited training data, poses challenges for generalization, as similar trait scores can manifest through diverse, context-dependent behaviors. In this work, we explore the predictive impact of the more granular hierarchical levels of the Big-Five Personality Model, facets and nuances, to enhance personality recognition from audiovisual interaction data. Using the UDIVA v0.5 dataset, we trained a transformer-based model including cross-modal (audiovisual) and cross-subject (dyad-aware) attention mechanisms. Results show that nuance-level models consistently outperform facet and trait-level models, reducing mean squared error by up to 74\% across interaction scenarios.
\end{abstract}

\begin{IEEEkeywords}
 Nonverbal signals, Machine learning, Multimodal recognition, Big-Five Personality Model, Nuance-level Model, Personality Computing
\end{IEEEkeywords}

\section{Introduction}
Personality refers to distinct cognitive and behavioral patterns that vary across individuals and are relatively stable over time \cite{funder2001personality}. Personality self-perception is typically obtained via self-reported questionnaires, often based on the Big Five Model \cite{goldberg1993structure}, which summarizes personality into five broad traits: Openness to Experience, Conscientiousness, Extraversion, Agreeableness, and Negative Emotionality. In recent years, automatic personality recognition has attracted increasing attention, particularly focusing on inferring such traits from audiovisual or textual data for improved psychology assessment and personalized human-machine interaction~\cite{vinciarelli2014survey, celli2025twenty}.
However, these trait scores often fail to reflect how behavior varies across individuals and contexts. People with similar scores may exhibit different behaviors depending on personal and situational factors. This variability makes it difficult for machine learning models to learn consistent patterns from trait-level labels, especially when training data is limited.

Instead, we argue that more fine-grained levels of personality should complement broad trait measures, enhancing both the understanding and detection of behavioral expressions. As such, the Big Five Model allows for a hierarchical representation of personality, moving from high-level traits to lower-level facets and nuances~\cite{goldberg1993structure}. Facets are the next hierarchical level, disaggregating general traits into more detailed and meaningful aspects (e.g., Negative Emotionality into Anxiety, Depression, and Emotional Volatility). They capture important personality variations that might go unnoticed when aggregating them to the trait level. For instance, divergent results in the facets of a trait can average into the same trait score~\cite{anglim2020predicting}. Nuances are the lowest level and are operationalized as the questionnaire’s items~\cite{stewart2021finer}. Despite the facets potential, nuances present an even higher capacity of bringing to the surface deeper and contextualized descriptions of individuals~\cite{mottus2017nuances}.  Previous research has confirmed the predictive power of nuances over traits, with~\cite{hall2020targeting} reporting small but consistent effects predicting personality using digital footprint data. However, their relevance remains unexplored in the context of nonverbal behavior from audiovisual data, which captures more expressive manifestations of personality.

This study investigates the predictability of each of the hierarchical levels, namely traits, facets, and nuances, using multimodal behavioral data. We evaluate model performance across these levels to examine whether shifting the level of analysis from traits to finer-grained labels improves the recognition of personality-related patterns across different settings. For this purpose, we use UDIVA v0.5~\cite{palmero2021context, palmero2022chalearn}, a dataset of face-to-face dyadic interactions involving both structured and unstructured tasks. This dataset enables the observation of individual personality expression across diverse contexts, while also capturing interactional cues from the interlocutor. Our approach builds on an interlocutor-aware, Transformer-based personality recognition model~\cite{dodd2023framework}, that incorporates information from action units, gaze, head pose, and audio from both participants to estimate personality of the target person.

\section{Methodology}
This section outlines the dataset, the automatic personality recognition methodology, and the procedure for aggregating nuances and facets back to their corresponding traits.
\subsection{UDIVA v0.5 Dataset}

Several publicly available datasets have been instrumental for audiovisual personality recognition~\cite{palmero2021context, celli2025twenty}. 
The ones that present dyadic or group interactions show that context, such as the behavior of the partner, provides important social cues that influence how personality traits manifest and are perceived, validating the need for context-aware personality modeling~\cite{palmero2021context}. This study is based on the UDIVA v0.5 dataset, featuring a collection of 80 hours of time-synchronized multimodal, multiview videos, in which 134 participants, distributed in 145 sessions, perform four different tasks in dyads \cite{palmero2021context, palmero2022chalearn}. The activities consisted of one free task, in which participants shared a conversation; and three structured tasks, composed of a competitive game (Ghost task), a cooperative game (Lego task), and a cognitive game (Animals task). The Talk task lasted around 5 minutes, whereas the duration of the other tasks depended on participant performance. The dataset is divided into per-task and per-dyad (interaction session) videos. Each participant was featured in a minimum of one and a maximum of five interaction sessions with different known and unknown partners, thus capturing a broader range of behavioral variability. In addition, participants answered a set of questionnaires, including the Big Five Inventory (BFI-2)~\cite{soto2017bfi2}, which consists of 60, 1-to-5 Likert-scale items designed to measure 15 facets. In other words, each facet is assessed by four items (i.e., nuances), and each trait by 12 items.

\subsection{Self-reported personality regression}
\begin{figure}[t!]
    \centering
    \includegraphics[width=0.49\textwidth]{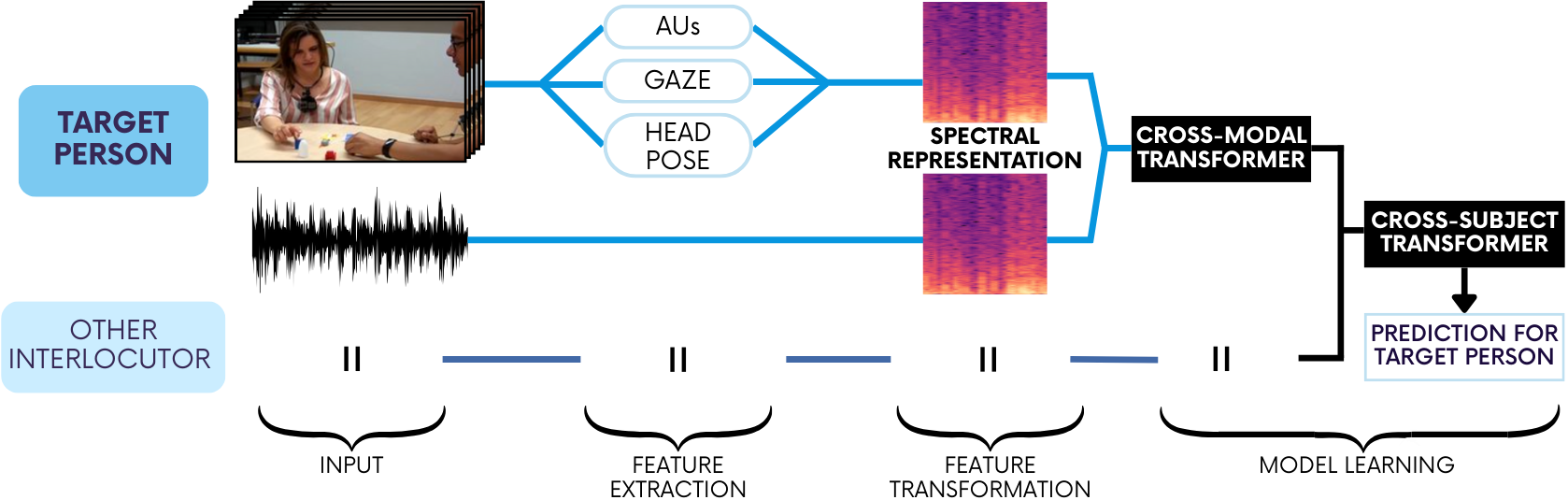} 
    \caption{Self-reported personality regression pipeline based on~\cite{dodd2023framework}.}
    \label{fig:Methodology}
\end{figure}

To achieve the goal of the study, we followed the interlocutor-aware, self-reported personality regression approach proposed by~\cite{dodd2023framework}, the current state-of-the-art on UDIVA v0.5. Unlike other approaches (e.g.,~\cite{palmero2021context, palmero2022chalearn, liao2024open}) that estimate personality from fixed-length video snippets and subsequently aggregate predictions for the given video, ~\cite{dodd2023framework} employs a spectral feature representation that leverages all available cues from an entire video, enabling more holistic and context-aware personality estimation. A diagram of the approach is shown in Fig.~\ref{fig:Methodology}.

In particular, we first extracted per-frame features from three visual modalities using OpenFace 2.0~\cite{baltrusaitis2018openface}: 35 facial action units, out of which 17 were intensity and the rest were presence values; gaze vectors for both eyes; and head position and orientation. Next, we converted these temporal signals as well as the audio signal into a spectral representation with the Discrete Fourier Transform. This transformation normalizes the input length and helps capture multi-scale temporal patterns. Each modality was transformed into a fixed-size 80-dimensional spectral map.

Finally, we fed these features into a multimodal Transformer architecture based on MulT~\cite{tsai2019multimodal}, to regress personality scores for each task video independently.
More specifically, each modality is processed via self-attention to capture intra-modal features, while cross-modal attention layers enable pairwise interaction between modalities, allowing one modality to attend to and reinforce another’s low-level features. %The architecture is composed of stacked bidirectional cross-modal attention blocks, which eliminates the need for strict temporal alignment between modalities. 
For multi-subject fusion, the architecture is scaled to include four cross-modal and self-attention blocks per participant, followed by a cross-subject attention layer to model interactions between participants. According to~\cite{dodd2023framework}, this approach outperforms traditional fusion methods (e.g., feature concatenation and averaging) by preserving modality-specific nuances while learning context-aware representations.

Using this approach, we compute personality scores of the target person at trait, facet, and nuance levels by training separate, level-specific models; that is, each using a distinct set of labels corresponding to its respective level.

\subsection{Score aggregation strategy}
The hierarchical structure of the Big Five Model allows for the aggregation of lower-order elements, such as facets and nuances, into higher-order traits. 
In our study, aggregation serves two purposes: (1) to unify all predictions at the trait level to enable fair comparison across trait, facet, and nuance models; and (2) to collapse repeated instances of the same participant in the test set due to multiple sessions into a single personality profile. 

For facet and nuance-level models, we first convert video-level predictions to the trait level by following the scoring book of the BFI-2 model~\cite{soto2017bfi2} before collapsing participant-level duplicates. The former involves computing the mean of the three relevant facet scores to form each trait score (for facets), and computing the mean of the 12 relevant item scores per trait (for nuances), while properly accounting for reverse-coded items. This transformation is mathematically equivalent to converting nuances into facets before being converted into traits.  Once nuances and facets are converted into traits, and also for the estimates of the trait-level model, we aggregate repeated per-person predictions into a single score by computing the mean.

\section{Experimental evaluation and results}
In this section, we describe the evaluation protocol and present the results obtained in our study. In particular, we aim to compare the predictive performance across the different hierarchy levels, and to assess whether personality traits can be better estimated from lower levels.

\subsection{Evaluation protocol}
We create 10-fold subject-independent training, validation, and test splits, aiming at a close-to-uniform distribution of personality trait scores, age, and gender among participants in the training set. Splits follow approximately an 8:1:1 ratio. 

We used Mean Squared Error (MSE) as training loss and Adam as optimizer. We optimized the model performance with Bayesian hyperparameter tuning using early stopping based on the validation loss. For each fold, 10 sweeps were performed to search for optimal learning rate and batch size. The optimal values for learning rate were found in the range of [0.01, 0.00001] and for batch size [16,32]. Final performance is evaluated and reported on each fold's test set.

We train one model per interaction task and personality level.
To assess the effectiveness of our models, we report results on multiple metrics: MSE, Mean Absolute Error (MAE), Pearson Correlation Coefficient (PCC), and R-squared (R$^2$). We also perform a cross-level comparison by aggregating predictions from lower levels (nuances and facets) and contrasting them with the higher level (trait). As baseline, we compute the average of per-trait (or per-facet, per-nuance) scores on the training set and apply them as predictions on the test set.

\subsection{Trait-level results}
Table~\ref{tab:aggregated_results} and Table~\ref{tab:nonaggregated_results} show the results for self-reported personality recognition across tasks, personality traits, and level-specific models, aggregated and reported at the trait level. We observe that the nuances models demonstrate consistent improvements across all combinations, achieving substantially lower MSE and MAE and higher PCC and R${^2}$ than baseline (up to 87.46\% MSE decrease), traits (73.89\%), and facets (63.47\%). Similarly, the facets models consistently outperform traits (up to 31.75\% decrease). Standard deviation values are generally low, suggesting consistent performance across folds. 

\subsection{Facet- and nuance-level results}

To assess the performance of the lower-level models on their respective personality labels, and to understand the effect of score aggregation, Table~\ref{tab:nuance_facet_results} reports the results for the nuance model at nuance level, and for the facet and nuance model at facet level, compared to their respective baselines. Again, all models significantly outperform the baseline. The observed improvement across all metrics when aggregating nuances into facets suggests that, although nuances may be more variable individually, their predictive reliability is enhanced when averaged into broader personality constructs.

Upon further analysis, we observe that the best-performing individual facets are: Respectfulness and Compassion, related to Agreeableness;  Energy Level, related to Extraversion; Creative Imagination, related to Openness; and Productiveness, related to Conscientiousness. However, Depression, Emotional Volatility, and Anxiety consistently performed worst overall, representing all facets of Negative Emotionality. 

\begin{table}[t!]
  \caption{Self-reported personality recognition results (MSE) across tasks. Each model is trained using different personality hierarchy levels (traits, facets, nuances), and evaluated at trait level. Results are averaged over 10 participant-independent test folds. Bold: best result per task. Underlined: best result overall.
  }
  \label{tab:aggregated_results}
  \centering
  \resizebox{\columnwidth}{!}{%
    \begin{tabular}{llcccccc}
      \toprule
      \textbf{Task} & \textbf{Model} & \textbf{O} & \textbf{C} & \textbf{E} & \textbf{A} & \textbf{N} & \textbf{Mean} %& \textbf{SD} 
      \\
      \midrule
      -
        & Baseline & 0.3363 & 0.4608 & 0.3064 & 0.2159 & 0.5179 & 0.3675 %& 0.1132 
        \\
      \midrule
      \multirow{4}{*}{Talk}
        & Traits   & 0.1643 & 0.1689 & 0.1618 & 0.1431 & 0.2323 & 0.1741 %& 0.0307 
        \\
        & Facets   & 0.1112 & 0.1193 & 0.1114 & 0.1005 & 0.1504 & 0.1185 %& 0.0170 
        \\
        & Nuances  & \textbf{0.0986} & \textbf{0.0496} & \textbf{0.0302} & \textbf{0.0400} & \textbf{0.0276} & \textbf{0.0492} %& \textbf{0.0258} 
        \\
      \midrule
      \multirow{3}{*}{Ghost}
        & Traits   & 0.1590 & 0.1738 & 0.1729 & 0.1318 & 0.2667 & 0.1808 %& 0.0464 
        \\
        & Facets   & 0.1266 & 0.1181 & 0.1100 & 0.1101 & 0.1521 & 0.1234 %& 0.0163
        \\
        & Nuances  & \textbf{0.0922} & \textbf{0.0509} & \textbf{0.0288} & \underline{\textbf{0.0381}} & \textbf{0.0261} & \textbf{0.0472} %& \textbf{0.0252} 
        \\
      \midrule
      \multirow{3}{*}{Lego}
        & Traits   & 0.1530 & 0.1734 & 0.1666 & 0.1327 & 0.2387 & 0.1729 %& 0.0386 
        \\
        & Facets   & 0.1221 & 0.1266 & 0.1122 & 0.1120 & 0.1582 & 0.1262 %& 0.0174 
        \\
        & Nuances  & \underline{\textbf{0.0913}} & \underline{\textbf{0.0478}} & \underline{\textbf{0.0278}} & \textbf{0.0382} & \underline{\textbf{0.0256}} & \underline{\textbf{0.0461}} %& \textbf{0.0250} 
        \\
      \midrule
      \multirow{3}{*}{Animals}
        & Traits   & 0.1547 & 0.1702 & 0.1706 & 0.1247 & 0.2407 & 0.1721 %& 0.0416 
        \\
        & Facets   & 0.1162 & 0.1163 & 0.1115 & 0.1033 & 0.1505 & 0.1196 %& 0.0163 
        \\
        & Nuances  & \textbf{0.0947} & \textbf{0.0493} & \textbf{0.0289} & \textbf{0.0389} & \textbf{0.0266} & \textbf{0.0477} %& \textbf{0.0246} 
        \\
      \bottomrule
    \end{tabular}
  }
\end{table}

\begin{table}[t!]
\caption{Self-reported personality recognition results across tasks. Each model is trained at different personality hierarchy levels and evaluated at trait level. Results are reported as the average ($\pm$ standard deviation) over 10 participant-independent test folds. Bold: best result per task. Underlined: best result overall.}
\label{tab:nonaggregated_results}
\centering
\scriptsize % Reduce font size to fit
\setlength{\tabcolsep}{4pt} % Reduce column padding
\begin{tabular}{llcccc}
\toprule
\textbf{Task} & \textbf{Model} & 
\textbf{MSE} $\downarrow$ & \textbf{MAE} $\downarrow$ & \textbf{PCC} $\uparrow$ &\textbf{R$^2$} $\uparrow$ \\
\midrule
- & Baseline & 0.3675\textsuperscript{$\pm$0.121}& 0.4826\textsuperscript{$\pm$0.091}& -& - \\
\midrule
\multirow{3}{*}{Talk} & Traits   & 0.1741\textsuperscript{$\pm$0.034}& 0.3327\textsuperscript{$\pm$0.333} & 0.7126\textsuperscript{$\pm$0.050}  & 0.3393\textsuperscript{$\pm$0.126}\\
     & Facets   & 0.1185\textsuperscript{$\pm$0.019}  & 0.2707\textsuperscript{$\pm$0.016}& 0.7960\textsuperscript{$\pm$0.061} & 0.5648\textsuperscript{$\pm$0.105} \\
     & Nuances  & \textbf{0.0492}\textsuperscript{$\pm$0.029}& \textbf{0.1628}\textsuperscript{$\pm$0.043}& \textbf{0.9484}\textsuperscript{$\pm$0.030}& \textbf{0.8232}\textsuperscript{$\pm$0.124}\\
\midrule
\multirow{3}{*}{Ghost} & Traits  & 0.1808\textsuperscript{$\pm$0.051} & 0.3379\textsuperscript{$\pm$0.040} & 0.7141\textsuperscript{$\pm$0.049} & 0.3215\textsuperscript{$\pm$0.099}  \\
       & Facets  & 0.1234\textsuperscript{$\pm$0.017}& 0.2779\textsuperscript{$\pm$0.018} & 0.7902\textsuperscript{$\pm$0.063} & 0.5467\textsuperscript{$\pm$0.120} \\
      & Nuances & \textbf{0.0472}\textsuperscript{$\pm$0.027}& \textbf{0.1611}\textsuperscript{$\pm$0.043}& \textbf{0.9479}\textsuperscript{$\pm$0.031}& \textbf{0.8306}\textsuperscript{$\pm$0.117}\\
\midrule
    \multirow{3}{*}{Lego} & Traits   & 0.1729\textsuperscript{$\pm$0.040}  & 0.3280\textsuperscript{$\pm$0.033}  & 0.7124\textsuperscript{$\pm$0.051} & 0.3453\textsuperscript{$\pm$0.116}\\
     & Facets   & 0.1262\textsuperscript{$\pm$0.019}  & 0.2816\textsuperscript{$\pm$0.017} & 0.7947\textsuperscript{$\pm$0.061}& 0.5387\textsuperscript{$\pm$0.114} \\ 
     & Nuances  & \textbf{\underline{0.0461}}\textsuperscript{$\pm$0.027}& \underline{\textbf{0.1589}}\textsuperscript{$\pm$0.042}&\underline{\textbf{0.9499}}\textsuperscript{$\pm$0.029} & \underline{\textbf{0.8346}}\textsuperscript{$\pm$0.114}\\ 
\midrule
        \multirow{3}{*}{Animals} & Traits  & 0.1721\textsuperscript{$\pm$0.043}& 0.3296\textsuperscript{$\pm$0.036} & 0.7105\textsuperscript{$\pm$0.054} & 0.3473\textsuperscript{$\pm$0.095}\\
        & Facets  & 0.1196\textsuperscript{$\pm$0.018}& 0.2732\textsuperscript{$\pm$0.017}& 0.7967\textsuperscript{$\pm$0.061} & 0.5601\textsuperscript{$\pm$0.112} \\
        & Nuances & \textbf{0.0477}\textsuperscript{$\pm$0.028}& \textbf{0.1610}\textsuperscript{$\pm$0.044}& \textbf{0.9480}\textsuperscript{$\pm$0.031}& \textbf{0.8286}\textsuperscript{$\pm$0.118}\\
\bottomrule
\end{tabular}
\end{table}

\subsection{Differences across interaction tasks}
At the task level, we observe minimal ($<$0.01) differences in predictability across all tasks and personality hierarchical levels. This aligns with the trait-level results reported by~\cite{dodd2023framework}, whose spectral-based methodology we adopt in this study. By contrast, previous fixed-length space-time-based approaches (e.g.~\cite{palmero2021context}) report substantial task-dependent variability, suggesting that the chosen representation significantly affects how task-specific information is captured.

\subsection{Differences across personality traits}
At the trait level, we observe that Agreeableness is the best predicted one with the facets model (Talk task, MSE = 0.1005), and Negative Emotionality with the nuances model (Lego task, MSE = 0.0256). Interestingly, the latter is the worst predicted trait for the traits, facets, and baseline models, marking the highest performance improvement across traits (up to 95\% MSE decrease compared to the baseline), suggesting that the estimation of specific behaviors, as powered by the nuances model, is particularly useful for such trait.

\subsection{Additional aggregation strategies}
We assessed additional aggregation mechanisms in preliminary experiments, including the use of median instead of mean, and aggregating the repeated per-person predictions for facets and nuances before converting them into traits, both yielding decreased performance compared to the results previously reported. For the latter, we hypothesize that since the lower hierarchy levels reflect more specific, observable behaviors during interactions, averaging across sessions may dilute these fine-grained behavioral signals. Preserving session-level information, instead, may help retain the richness of both nuanced and facet-level expressions.

\begin{table}[t!]
\caption{Self-reported personality recognition results across tasks for nuance- and facet-specific models, aggregated and reported at different levels, %Facet-level models were evaluated at the facet level but trained at its corresponding hierarchy level and nuance-level models at the nuance level.
as the average ($\pm$ standard deviation) over 10 participant-independent test folds. %Baseline MSE/MAE values correspond to the mean predictor for each task.
Bold: best result per task. Underlined: best result overall.}
\label{tab:nuance_facet_results}
\centering
\scriptsize
\setlength{\tabcolsep}{4pt}
\begin{tabular}{llcccc}
\toprule
\textbf{Task} & \textbf{Model} & 
\textbf{MSE} $\downarrow$ & \textbf{MAE} $\downarrow$ & \textbf{PCC} $\uparrow$ &\textbf{R$^2$} $\uparrow$ \\
\midrule
\multicolumn{6}{c}{\textbf{Nuance Level}} \\
\midrule
- & Baseline & 1.0676\textsuperscript{$\pm$0.319} & 0.8362\textsuperscript{$\pm$0.184} & - & - \\ 
\midrule
Talk    & Nuances & 0.2075\textsuperscript{$\pm$0.091}& 0.3507\textsuperscript{$\pm$0.082}& \underline{0.9185}\textsuperscript{$\pm$0.057}& 0.6636\textsuperscript{$\pm$0.232}\\
Ghost   & Nuances & 0.2092\textsuperscript{$\pm$0.087}& 0.3544\textsuperscript{$\pm$0.079}& 0.9161\textsuperscript{$\pm$0.059}& 0.6590\textsuperscript{$\pm$0.233}\\
Lego    & Nuances & \underline{0.2051}\textsuperscript{$\pm$0.085}& 0.3483\textsuperscript{$\pm$0.081}& 0.9182\textsuperscript{$\pm$0.058}& 0.6638\textsuperscript{$\pm$0.234}\\
Animals & Nuances & 0.2067\textsuperscript{$\pm$0.088}& \underline{0.3457}\textsuperscript{$\pm$0.080}& 0.9179\textsuperscript{$\pm$0.058}& \underline{0.6651}\textsuperscript{$\pm$0.230}\\ 
\midrule
\multicolumn{6}{c}{\textbf{Facet Level}} \\
\midrule
- & Baseline & 0.5416\textsuperscript{$\pm$0.184}& 0.5975\textsuperscript{$\pm$0.115}& - & - \\
\midrule
\multirow{2}{*}{Talk}    
    & Facets  & 0.1631\textsuperscript{$\pm$0.031}& 0.3208\textsuperscript{$\pm$0.028}& 0.8277\textsuperscript{$\pm$0.043}& 0.5940\textsuperscript{$\pm$0.108}\\
    & Nuances & \textbf{0.0777\textsuperscript{$\pm$0.034}}& \textbf{0.2089\textsuperscript{$\pm$0.057}}& \textbf{0.9446\textsuperscript{$\pm$0.041}}& \textbf{0.8072\textsuperscript{$\pm$0.127}}\\

\midrule
\multirow{2}{*}{Ghost}
    & Facets  & 0.1681\textsuperscript{$\pm$0.029}& 0.3243\textsuperscript{$\pm$0.026}& 0.8231\textsuperscript{$\pm$0.048}& 0.5831\textsuperscript{$\pm$0.119}\\
    & Nuances & \textbf{0.0770\textsuperscript{$\pm$0.033}}& \textbf{0.2093\textsuperscript{$\pm$0.057}}& \textbf{0.9435\textsuperscript{$\pm$0.041}}& \textbf{0.8060\textsuperscript{$\pm$0.129}}\\

\midrule
\multirow{2}{*}{Lego} 
    & Facets& 0.1728\textsuperscript{$\pm$0.030}& 0.3286\textsuperscript{$\pm$0.026}& 0.8243\textsuperscript{$\pm$0.045}& 0.5754\textsuperscript{$\pm$0.114}\\
    & Nuances & \textbf{\underline{0.0749}\textsuperscript{$\pm$0.033}}& \textbf{\underline{0.2066}\textsuperscript{$\pm$0.057}}& \textbf{\underline{0.9456}\textsuperscript{$\pm$0.039}}& \textbf{\underline{0.8099}\textsuperscript{$\pm$0.129}}\\

\midrule
\multirow{2}{*}{Animals} 
    & Facets  & 0.1634\textsuperscript{$\pm$0.030}& 0.3193\textsuperscript{$\pm$0.027}& 0.8279\textsuperscript{$\pm$0.046}& 0.5969\textsuperscript{$\pm$0.108}\\
    & Nuances & \textbf{0.0765\textsuperscript{$\pm$0.034}}& \textbf{0.2075\textsuperscript{$\pm$0.057}}& \textbf{0.9443\textsuperscript{$\pm$0.040}}& \textbf{0.8081\textsuperscript{$\pm$0.127}}\\

\bottomrule
\end{tabular}
\end{table}

\section{Conclusions and future work}
In this paper, we leveraged the hierarchy of the Big Five model, consisting of nuances, facets, and traits, to assess whether predicting lower-level constructs improves self-reported personality recognition. Results showed that nuance-level labels significantly improve accuracy on audiovisual dyadic data, highlighting their potential for more precise and context-sensitive personality modeling. 

Future work will analyze individual nuance and facet performance, explore links between audiovisual features and top-performing nuances, and examine the effects of demographic bias. We ultimately plan to expand the study to additional methods and datasets with raw scores for a deeper evaluation of personality, enabling a more comprehensive evaluation of whether, where traits see a forest, nuances capture every leaf.

\section*{Ethical Impact Statement}
We recognize that biased affective computing models from non-diverse data (e.g., lacking cultural, age, and relational range) lead to significant ethical risks. In applying personality computing models to employment and health contexts, this bias can manifest itself as harmful stereotyping, ultimately leading to discriminatory decisions. Future studies must consider these limitations when designing new applications.

\section*{Acknowledgments}
This work has been partially supported by the Spanish project PID2022-136436NB-I00, ICREA Academia program, and EU Erasmus Mundus Joint Master Grant no. 101048710. We thank Albert Clapés for creating the dataset split.

\bibliographystyle{ieeetr}
\bibliography{./sample-base2}

\end{document}